\documentclass[11pt]{article}

\usepackage[utf8]{inputenc}
\usepackage[T1]{fontenc}
\usepackage{lmodern}
\usepackage{amsmath}
\usepackage{amssymb}
\usepackage{array}
\usepackage{booktabs}
\usepackage{geometry}
\usepackage[authoryear,round]{natbib}
\usepackage{hyperref}
\usepackage{microtype}
\usepackage{graphicx}
\usepackage[section]{placeins}

\hypersetup{hidelinks}
\newcolumntype{L}[1]{>{\raggedright\arraybackslash}p{#1}}

\title{How Much Dense Attention is Necessary?\\
Oracle-Guided Sparse Prefill for Full/GQA Layers in Hybrid Long-Context Models}

\author{
\begin{tabular}{ccc}
Hongxing WANG & Harenome RAZANAJATO & Zhen ZHANG \\[0.6em]
\multicolumn{3}{c}{Yujie YUAN \qquad Hongsheng LIU}
\end{tabular}\\[0.5em]
}

\date{Technical Report, First Release}

\begin{document}

\maketitle

\begin{abstract}
Long-context prefill remains expensive because full or grouped-query attention (GQA) layers still score the historical sequence, even in models that also use local, sparse, linear, or hybrid components. This technical report asks a diagnostic question for sparse prefill: under an explicit support granularity and top-$k$ budget, how much dense attention is needed to preserve task-level behavior?

We introduce an attention-mass top-$k$ oracle for existing GQA checkpoints. For each layer and query position, the oracle computes dense attention, selects a head-averaged token support, and recomputes attention restricted to that support. The oracle is not a deployable acceleration method; it is a controlled reference for separating sparse-budget feasibility from indexer error and runtime realization effects.

Across the reported Qwen-family retrieval-heavy evaluations, the longest per-query oracle rows stay within 1 point of dense, and a Qwen3.5-9B RULER-style sweep from 4K to 100K stays within 0.48 points. Guided by this oracle, we derive a head-collapsed auxiliary indexer whose supervision is the oracle's dense attention-mass distribution, training it by KL distillation from dense attention while keeping the backbone frozen. With separately distilled Qwen3.5-0.8B and Qwen3.5-9B indexers, the reported 16K/32K validation macro gaps are +2.04 and +1.13 points, treated as quality preservation rather than improvement; the fused selection-block-shared path can introduce a larger realization gap.

We also report preliminary single-card TTFT measurements on both NPU and GPU. With distilled indexer weights, sparse serving reaches 1.71$\times$ for Qwen3.5-0.8B on NPU against the vLLM Ascend FlashAttention baseline and 1.93$\times$ for Qwen3.5-9B on GPU against its dense FlashAttention-2 baseline. Additional stress tests that run the same sparse serving path with untrained indexer weights reach 3.44$\times$, indicating sparse-runtime headroom but not validated output quality. This first release therefore separates oracle feasibility, distilled-indexer quality, and sparse-runtime headroom, while leaving a fully matched dense--oracle--distilled quality-latency frontier to future work.
\end{abstract}

\section{Introduction}

Long-context prefill is a major cost in serving language models because the prompt must be processed before the first generated token. Modern long-context architectures increasingly combine local, sparse, linear, recurrent, and full-attention components, as in Gemma-style hybrids, Qwen3-Next, Jamba, and related deployments \citep{gemma2024,gemma2025,qwen3next2025,jamba2024,mistral2023}. Full or grouped-query attention (GQA) layers often remain the places where long-range evidence is retrieved and fused. GQA reduces KV-cache size by sharing key-value heads \citep{ainslie2023gqa}, but it does not remove the dense search over historical tokens inside full-attention layers. This raises a diagnostic question for sparse prefill: \emph{under a fixed support granularity and top-$k$ budget, how much of this dense attention search is actually needed to preserve task-level behavior?}

Sparse attention is a natural way to reduce this search: if each query only needs a small subset of context tokens, full attention can be replaced by query-dependent top-$k$ attention over selected KV positions. Prior systems such as Quest~\citep{tang2024quest}, DuoAttention~\citep{xiao2024duoattention}, and DeepSeek Sparse Attention~\citep{deepseek2025dsa} provide evidence that query-aware or learned token selection can preserve quality while reducing long-context cost under their evaluated settings. Dynamic sparse-prefill systems such as MInference~\citep{jiang2024minference} and native sparse-attention architectures such as NSA~\citep{yuan2025nsa} address related systems goals from different design points. However, a sparse-prefill result is hard to interpret without a reference: a quality drop could come from an insufficient budget, a distilled indexer that misses the right support, support sharing that is too coarse for the task, or implementation-level differences in the runtime path.

This paper takes an oracle-first view of that problem. We introduce an \emph{attention-mass top-$k$ oracle}: for each layer and query position, the oracle computes dense attention, selects a head-averaged top-$k$ token support under a specified support granularity, and then recomputes attention restricted to that support. The oracle is not a deployable acceleration method; its role is to provide a controlled reference that separates whether a sparse budget and support granularity are plausible from whether an inference-time indexer and runtime realization can approximate them.

\begin{figure}[htbp]
    \centering
    \includegraphics[width=0.9\textwidth]{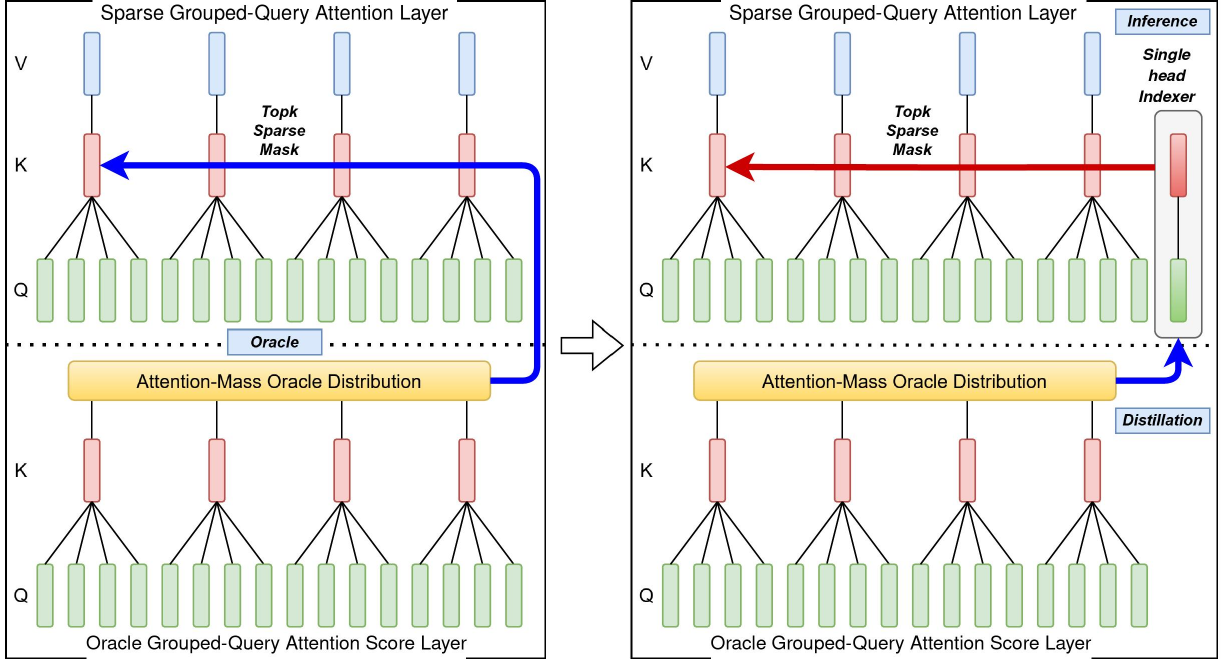}
    \caption{Oracle-guided sparse prefill workflow. Dense attention mass supplies a top-$k$ support reference for oracle evaluation and a teacher signal for indexer distillation.}
    \label{fig:oracle-workflow}
\end{figure}

Figure~\ref{fig:oracle-workflow} summarizes this oracle-guided workflow. Dense attention first defines a support reference; the same support signal is then used to guide budget and granularity choices, train an indexer, and evaluate the gap between oracle support, distilled-indexer support, and runtime support sharing.

The reported oracle evidence supports the usefulness of this reference in Qwen-family GQA checkpoints. On non-public RULER-style summaries, per-query dense-attention top-$k$ remains near dense at the longest evaluated lengths, including 128K rows for Qwen3.5/Qwen3.6 27B and 35B-A3B checkpoints and a 32K row for Qwen3-8B. A separate Qwen3.5-9B RULER-13 oracle-prior sweep covers 4K--100K with 3900 samples and approximately 140M tokens under a configured power-of-two top-$k$ schedule from 256 to 2048, with every oracle-minus-dense gap within 0.48 points. These results suggest that, for the evaluated retrieval-heavy slices, dense attention mass often identifies a compact support that preserves task-level behavior. They do not imply that the selected support is the best possible sparse support for the downstream task.

Guided by the oracle, we derive a head-collapsed indexer design for frozen GQA backbones. The original model remains unchanged. The indexer learns to predict token importance by KL distillation from the same dense attention-mass distributions used by the oracle, collapses the auxiliary scoring space across heads, and leaves the sparse top-$k$ budget adjustable at inference time. This design follows from the oracle-guided argument: if oracle attention-mass sparse support is sufficient for a task slice, then the inference-time module should focus on approximating that support with limited additional machinery instead of modifying the backbone.

This first release reports three sparse evidence types and keeps their boundaries explicit. Oracle sparse rows use dense attention to compute sparse support and test whether the evaluated support family can preserve quality. Distilled-indexer sparse rows use real distilled Qwen3.5 indexer weights and provide the main real-indexer evidence for frozen-backbone sparse prefill with dense decode and TTFT speedup. Additional stress rows use untrained indexer weights only to exercise the same sparse serving path and expose runtime headroom. For checkpoints where oracle rows are also available, the oracle and stress rows are complementary evidence for sparse feasibility and systems scaling, not a single validated deployment configuration. We therefore treat this version as a scoped technical report and leave the fully matched dense--oracle--distilled quality-latency frontier to future work.

Our contributions are:
\begin{itemize}
\item We formulate sparse prefill for full/GQA attention layers as an oracle-guided reducibility question under explicit top-$k$ and support-granularity constraints.
\item We define an attention-mass top-$k$ oracle that selects head-averaged token support from dense attention and use it to organize evidence around oracle gap, indexer gap, and realization gap.
\item We report first-release oracle evidence on Qwen-family long-context benchmark slices, showing that the evaluated attention-mass supports often remain near dense under the reported budgets.
\item We derive and evaluate a frozen-backbone, head-collapsed, KL-distilled indexer that predicts oracle-style token importance while leaving the sparse budget adjustable at inference time.
\item We report distilled-indexer quality measurements and single-card distilled-indexer TTFT measurements, while using additional untrained-indexer stress rows to characterize sparse-runtime headroom rather than matched output-quality evidence.
\end{itemize}

\section{Background and Motivation}

\subsection{Hybrid Long-Context Architectures}

The design space of long-context language models is moving away from uniformly dense Transformer attention. Many recent architectures combine multiple sequence-processing mechanisms: local windows handle nearby context, global or full attention layers preserve long-range routing, linear recurrent components improve length scaling, and sparse modules reduce the amount of KV state touched by each query \citep{gemma2024,gemma2025,qwen3next2025,jamba2024,deepseek2025dsa}. This broad shift reflects a practical fact: full dense attention is powerful, but using it everywhere becomes increasingly expensive as the context length grows.

The implication for sparse prefill is subtle. Hybrid architectures do not make full attention irrelevant. Instead, they make the remaining full or GQA attention layers more important as quality-critical global communication points. These layers are natural targets for retrofit sparse prefill: replacing them with sparse computation can reduce TTFT, but only if the selected support preserves the information that full attention would have retrieved.

\subsection{GQA and the Remaining Dense Search Cost}

Grouped-query attention reduces the number of KV heads by allowing several query heads to share a KV head \citep{ainslie2023gqa}. This lowers KV-cache memory and bandwidth, which is essential in long-context serving. However, for a full-attention layer with sequence length $S$, each query still scores the historical KV sequence. GQA changes the head structure and cache footprint, but the query-dependent search space remains dense in the sequence dimension.

This distinction motivates our focus. We do not aim to redesign the backbone attention architecture. We ask whether the dense sequence search inside existing full/GQA attention layers can be replaced by a sparse, query-dependent support while keeping the original checkpoint fixed.

\subsection{Learned Sparse Attention and DSA}

Data-dependent sparse attention methods show that the important context tokens differ by query. Fixed sparse patterns are useful for some settings \citep{beltagy2020longformer,zaheer2020bigbird}, but they do not directly adapt to input-dependent retrieval locations or task-dependent evidence. Learned indexers go further by predicting which KV positions should be retained for each query \citep{tang2024quest,xiao2024duoattention,deepseek2025dsa}. DSA is a particularly important reference point because it demonstrates that an indexer can support efficient long-context sparse attention when the architecture and serving stack are designed around that goal. Dynamic sparse-prefill methods such as MInference exploit structured attention patterns at inference time without fine-tuning, while native sparse-attention architectures such as NSA train sparsity into the model itself \citep{jiang2024minference,yuan2025nsa}. Our setting sits between these lines: the backbone is frozen, but the support predictor is learned by distillation from the dense attention distribution rather than selected by a fixed or purely heuristic rule.

Our work asks a complementary question. Instead of starting with a deployable sparse system, we first measure the sparse support induced by dense attention mass under explicit budget and granularity constraints. This oracle-first methodology clarifies whether errors come from the sparse budget, from the distilled indexer, or from the runtime realization.

\subsection{Why an Oracle is Needed}

Without an oracle, sparse-prefill results are difficult to interpret. A quality drop could mean that top-$k$ sparse attention is inherently insufficient, that the inference-time indexer selected the wrong support, that support sharing was too coarse, or that the implementation introduced numerical or kernel-level differences. An oracle does not solve deployment, but it makes the scientific question more interpretable. It tells us whether the dense-attention-mass support under the same budget and granularity would have preserved quality.

In this paper the oracle also guides the method. First, it defines the target distribution that the indexer should learn. Second, it reveals which top-$k$ budgets are plausible at different context lengths. Third, it compares support granularities, such as per-query and selection-block-shared support, before committing to a systems realization. Fourth, it supplies the reference points used to separate oracle gap, indexer gap, and realization gap. Thus the oracle is not a post-hoc visualization tool; it is the mechanism that turns dense attention behavior into a sparse-prefill design.

\subsection{Positioning Relative to Prior Work}

Fixed sparse-attention patterns such as Longformer and BigBird reduce asymptotic cost by imposing local, global, or random connectivity patterns \citep{beltagy2020longformer,zaheer2020bigbird}. These patterns are well matched to fixed sparse schedules, but they are not query-adaptive and are not designed to react to input-dependent evidence locations in retrieval-heavy prompts. Our oracle instead asks whether the dense attention distribution of an existing checkpoint already concentrates on a compact, query-dependent support.

Retrieval- or streaming-head methods such as Quest and DuoAttention identify attention structures that can be served more cheaply \citep{tang2024quest,xiao2024duoattention}. Dynamic sparse-prefill methods such as MInference are especially close to the systems goal because they target TTFT with sparse prefill, but they do not use an oracle-distilled indexer or organize evaluation around the reducibility gap \citep{jiang2024minference}. Learned indexer systems such as DSA go further by training an indexer for sparse long-context inference, while NSA represents a native trainable-sparsity direction \citep{deepseek2025dsa,yuan2025nsa}. Our indexer is intentionally in this learned-indexer family rather than a claim of a fundamentally new sparse-attention primitive. The distinction is in emphasis: the backbone is frozen, the training target is the oracle's head-averaged dense attention-mass distribution, the sparse budget remains controllable at inference time, and the evaluation is organized around oracle gap, indexer gap, and realization gap.

Kernel-level methods such as FlashAttention and FlashAttention-2 improve exact dense attention without changing the attention support \citep{dao2022flashattention,dao2023flashattention2}. They are therefore complementary baselines: they reduce memory traffic for dense attention, while our sparse prefill changes which KV positions are read during full/GQA attention layers. Hybrid architectures such as Gemma, Jamba, and Qwen3-Next combine local, global, recurrent, linear, and full-attention components \citep{gemma2024,gemma2025,jamba2024,qwen3next2025}. The method and problem formulation apply to GQA layers generally; this first release focuses its empirical evidence on Qwen-family GQA checkpoints.

Our contribution is orthogonal to these categories. We do not begin from a standalone sparse-kernel or indexer-architecture claim. We first measure the reducibility of dense attention under explicit budget and support-granularity constraints, then use the oracle to guide indexer distillation, runtime granularity, and evaluation. This oracle-guided framing is the main distinction from systems that only report the final sparse implementation result.

\section{Problem Formulation}
\label{sec:problem}

Consider a frozen causal language model with a full-attention GQA layer $\ell$. Let $H_q$ be the number of query heads, $H_{kv}$ the number of KV heads, and $g:\{1,\ldots,H_q\}\rightarrow\{1,\ldots,H_{kv}\}$ the GQA group mapping from query head to KV head. For query position $t$ and key position $s\leq t$, dense GQA attention computes
\[
A_{\ell,t,h,s}
= \mathrm{softmax}_{s'\leq t}
\left(
\frac{
\langle q_{\ell,t,h}, k_{\ell,s',g(h)}\rangle
}{\sqrt{d_h}}
\right)_s,
\qquad
o_{\ell,t,h}
= \sum_{s\leq t} A_{\ell,t,h,s} v_{\ell,s,g(h)} .
\]
Thus GQA reduces KV-head multiplicity but does not remove the dense search over the sequence dimension.

Sparse prefill replaces the full historical sequence with a support set. In this paper, support granularity has two distinct axes:
\begin{itemize}
\item \textbf{Head axis:} support may be selected separately per query head $\mathcal{S}_{\ell,t,h}$, shared within each KV group $\mathcal{S}_{\ell,t,r}$ for query heads with $g(h)=r$, or collapsed into one head-averaged token support $\mathcal{S}_{\ell,t}$ shared by all query heads at position $t$.
\item \textbf{Query-position axis:} support may be selected per query position ($b_{\mathrm{sel}}=1$), shared by a selection block of query positions ($b_{\mathrm{sel}}>1$), or selected at a coarser token/block level depending on the runtime.
\end{itemize}
Unless otherwise stated, the reported oracle and indexer experiments use head-averaged token support on the head axis. We denote the number of consecutive query positions sharing one selected support by $b_{\mathrm{sel}}$, the \emph{selection block size}. When $b_{\mathrm{sel}}=1$, we call the realization \emph{per-query}; when $b_{\mathrm{sel}}=64$, we call it \emph{selection-block-shared}. This notation separates query-position support sharing from the GQA head mapping $g(h)$. The head-averaged support choice is a deliberate systems-oriented design point, not a claim that head-specific evidence is always redundant.

For a head-averaged token support $\mathcal{S}_{\ell,t}$ with $|\mathcal{S}_{\ell,t}|\leq k$, sparse attention recomputes the original attention operation over selected keys:
\[
A_{\ell,t,h,s}^{\mathcal{S}}
= \mathrm{softmax}_{s'\in\mathcal{S}_{\ell,t}}
\left(
\frac{
\langle q_{\ell,t,h}, k_{\ell,s',g(h)}\rangle
}{\sqrt{d_h}}
\right)_s,
\qquad
o_{\ell,t,h}^{\mathcal{S}}
= \sum_{s\in\mathcal{S}_{\ell,t}} A_{\ell,t,h,s}^{\mathcal{S}} v_{\ell,s,g(h)} .
\]

The key question is how to choose the support. We separate three sources of error:
\begin{itemize}
\item \textbf{Oracle gap:} the gap between dense attention and oracle sparse attention when the support is selected by the attention-mass oracle at the same budget and granularity.
\item \textbf{Indexer gap:} the additional gap caused by replacing oracle-selected support with an inference-time indexer.
\item \textbf{Realization gap:} the additional gap introduced by runtime constraints such as selection-block-shared support, KV-group expansion, kernel fusion, and amortized top-$k$ selection.
\end{itemize}

This decomposition is the organizing principle of the paper. It also defines the workflow. First, the oracle measures whether the chosen support family is viable. Second, the oracle distribution $P_{\ell,t}$ becomes the distillation target for the indexer. Third, oracle sweeps over $k$ and support sharing choose the runtime operating region. Fourth, distilled and fused implementations are judged by how much additional indexer and realization gap they introduce relative to the oracle references.

\section{Attention-Mass Top-\texorpdfstring{$k$}{k} Oracle}

\subsection{Oracle Definition}

Let $A_{\ell,t,h,s}$ be the dense attention weight for layer $\ell$, query position $t$, query head $h$, and key position $s$. The oracle used in this release aggregates dense attention into a head-averaged token-importance distribution:
\[
P_{\ell,t}(s)
= \frac{1}{H_q}\sum_{h=1}^{H_q} A_{\ell,t,h,s}.
\]
The denominator in the equivalent normalized form is $H_q$ because each head's causal attention distribution sums to one over valid keys. The oracle support is
\[
\mathcal{S}^{\star}_{\ell,t}
= \mathrm{TopK}_s(P_{\ell,t}(s), k).
\]
Sparse attention is then recomputed over $\mathcal{S}^{\star}_{\ell,t}$ using the original model's queries, keys, and values. The oracle therefore uses dense attention only to choose the support, not to replace the sparse attention output.

In the full-model HF oracle evaluation, patched full-attention layers are executed recursively along the sparse trajectory. At each patched layer, the oracle computes dense attention scores from the current hidden states to choose the top-$k$ support, but the layer returns the recomputed sparse attention output to the model. Consequently, later oracle decisions are based on hidden states already affected by earlier sparse replacements rather than on a separate dense-teacher trajectory. Single-layer diagnostic utilities may also materialize dense attention outputs for comparison, but those dense outputs are not fed back into the full-model oracle path.

This oracle is intentionally narrower than a global sparse optimum. It evaluates whether head-averaged token support selected by dense attention mass is sufficient at a given budget. A per-head oracle or per-KV-group oracle could preserve head-specific evidence differently, and a task-optimized sparse support could in principle differ from the attention-mass support.

The oracle therefore relies on a specific empirical assumption: for the evaluated support family, task-relevant evidence is sufficiently reflected in the head-averaged dense attention-mass distribution. The oracle tables test this assumption for the reported model and task slices, but they do not make it a theorem or rule out better per-head, per-KV-group, or task-optimized sparse supports.

This first release uses the head-averaged token oracle throughout. In the terminology of this report, head-averaged token support means that selected token positions are shared across query heads at a query position. The report does not include per-head or per-KV-group oracle variants, retained attention mass, or output-level divergence. We therefore interpret the oracle as a controlled design reference for the reported support family, not as evidence that every possible head-specific sparse realization would behave the same way.

\begin{center}
\fbox{\begin{minipage}{0.94\linewidth}
\small
\textbf{Algorithm 1: Attention-mass top-$k$ oracle for one patched layer.}
\begin{enumerate}
\item Given current sparse-trajectory hidden states, form the layer's original GQA $Q,K,V$ tensors.
\item Compute dense causal attention scores and softmax normalizers tile by tile.
\item Reconstruct dense attention probabilities tile by tile and average them over query heads to obtain $P_{\ell,t}(s)$.
\item Select top-$k$ token indices from $P_{\ell,t}$ for $b_{\mathrm{sel}}=1$, or from a max aggregate of $P_{\ell,t}$ over the queries in each selection block when $b_{\mathrm{sel}}>1$.
\item Broadcast the selected token indices across KV heads, pad invalid causal slots with a sentinel, and recompute the original GQA attention only on the selected support.
\item Return the recomputed sparse attention output to the model, so later oracle decisions are made on the sparse trajectory.
\end{enumerate}
\end{minipage}}
\end{center}

\subsection{Per-Query and Selection-Block-Shared Oracles}

The per-query oracle selects one support set for each query position under the chosen head-axis and query-position-axis support granularity. It measures the quality of the attention-mass support at fine query granularity. A selection-block-shared oracle selects a support set for a block of $b_{\mathrm{sel}}$ consecutive queries and reuses it across that block. This better matches fused serving implementations but can introduce quality loss when different queries require different evidence tokens.

For a selection block $B$ of consecutive query positions, the reported selection-block-shared implementation aggregates valid per-query token scores with a max rule,
\[
P_{\ell,B}(s)=\max_{t\in B:\,s\leq t}P_{\ell,t}(s),
\qquad
\mathcal{S}^{\star}_{\ell,B}=\mathrm{TopK}_s(P_{\ell,B}(s),k).
\]
Causally invalid positions are masked before top-$k$ selection, and the resulting support is reused by all queries in $B$. During sparse recomputation, each query still applies its own causal validity mask, so support entries with $s>t$ are ignored for that query.

The distinction is important. A strong per-query oracle with a weaker selection-block-shared oracle implies that sparse attention is feasible, but the systems realization must preserve enough query specificity.

\subsection{Oracle as a Scientific Instrument and Design Guide}

The oracle is not an acceleration method because computing $P_{\ell,t}$ requires dense attention. Its first purpose is to act as a scientific instrument. It answers three questions:
\begin{itemize}
\item Is dense attention mass sufficiently concentrated for top-$k$ support to preserve quality?
\item How large does $k$ need to be at different context lengths?
\item How much of the distilled-indexer error is due to support prediction rather than the sparse budget itself?
\end{itemize}

Its second purpose is to guide the inference-time method. The oracle distribution $P_{\ell,t}$ becomes the KL teacher for the indexer. Oracle sweeps over $k$ and support sharing define the operating region where sparse prefill is likely to preserve quality. The per-query oracle identifies the reducibility of dense attention, while the selection-block-shared oracle estimates the cost of systems-friendly amortization. The distilled indexer is therefore not an arbitrary sparse module trained in isolation; it is a compressed predictor of the oracle-selected support under a serving-oriented cost model.

\section{Oracle Study}

\subsection{Experimental Scope}

The oracle study uses non-public RULER-style long-context evaluation summaries~\citep{hsieh2024ruler}. These summaries evaluate dense-attention top-$k$ oracles on Qwen3-8B and on Qwen3.5/Qwen3.6 27B and 35B-A3B checkpoints. Model names in this report are checkpoint identifiers. Qwen3.5 and Qwen3.6 denote internal checkpoint generations; 27B and 35B-A3B are model-size labels used in those identifiers, and A3B should be read as an active-parameter label for the corresponding MoE-style checkpoint rather than as an independently audited public parameter count. Because exact public revisions and full architecture metadata are not available for every row, comparisons across these labels should be interpreted as within-row dense-versus-sparse evaluations rather than controlled model-lineage or model-size comparisons.

The evaluated contexts cover 32K for Qwen3-8B and 32K, 64K, and 128K for the larger Qwen3.5/Qwen3.6 checkpoints. We also include a Qwen3.5-9B RULER-13 oracle-prior sweep covering 4K, 8K, 16K, 32K, 64K, and 100K. The configured top-$k$ schedule for this sweep is 256, 512, 1024, 2048, 2048, and 2048 across those lengths; sparsity values are rounded and use the causal-position accounting described below. We use this sweep as oracle feasibility and budget-schedule evidence rather than as a strict matched frontier row. Each RULER-style cell reports an average score on a 0--100 scale over retrieval-heavy long-context tasks, including code-word extraction (CWE), frequent-word extraction (FWE), needle-in-a-haystack (NIAH) variants, question answering (QA), and variable tracking (VT). We call these evaluations RULER-style because they follow RULER task families, while the raw prompts, scoring-script revisions, and model revision hashes are not included in this release.

We keep two oracle realizations separate. The per-query oracle has selection block size $b_{\mathrm{sel}}=1$, where each query selects its own support. The selection-block-shared oracle uses $b_{\mathrm{sel}}=64$, where one selected support is amortized across a group of 64 query positions. This selection-block notation is unrelated to the GQA query-head to KV-head mapping $g(h)$ in Section~\ref{sec:problem}. The per-query oracle estimates sparse reducibility under fine-grained attention-mass support selection. The selection-block-shared oracle estimates how much of that reducibility survives a more systems-friendly support-sharing policy.

The main metrics are dense-relative task score, oracle gap in percentage points, configured top-$k$ budget, sparsity under causal-position accounting, and sensitivity to support granularity. For a sequence of length $S$ and a top-$k$ token support, causal dense attention has $S(S+1)/2$ valid positions, while the top-$k$ support retains $kS-k(k-1)/2$ positions when $k\leq S$; the reported sparsity is therefore a causal token-pair fraction rather than a FLOP, memory, or latency fraction. For retrieval-heavy benchmarks, task-wise breakdowns are important because failures often concentrate in fine-grained token retrieval rather than broad semantic aggregation.

This technical report reports aggregate cells and the task families they cover, but it does not yet include a public artifact mapping each cell to raw prompts, model revisions, and scoring scripts. Positive oracle-minus-dense gaps are therefore treated as quality preservation within evaluation resolution rather than evidence that oracle sparse attention is intrinsically better than dense attention.

\subsection{Per-Query Oracle Feasibility}

Table~\ref{tab:oracle-longest} reports the longest reported oracle slice for each model under top-$k=2048$. This is not the smallest usable budget for every model; rather, it is a common high-quality budget across the reported summaries that tests whether the head-averaged attention-mass support preserves dense behavior at long context. The result is consistently near dense for the larger Qwen3.5/Qwen3.6 checkpoints, with gaps from +0.21 to -0.67 percentage points in the per-query oracle. Qwen3-8B is more sensitive, but still remains within 0.92 points at 32K.

\begin{table}[t]
\caption{Attention-mass oracle quality at the longest evaluated context for each model. Values are 13-task averages. Gap is oracle minus dense, in percentage points; small positive gaps are treated as evaluation variation, not as evidence that oracle sparse attention dominates dense attention.}
\label{tab:oracle-longest}
\centering
\small
\begin{tabular}{lrrrrr}
\toprule
Model & Context & Dense & Per-query top-$k$ & Per-query oracle & Gap \\
\midrule
Qwen3-8B & 32K & 90.17 & 2048 & 89.25 & -0.92 \\
Qwen3.5-27B & 128K & 96.46 & 2048 & 96.67 & +0.21 \\
Qwen3.5-35B-A3B & 128K & 96.72 & 2048 & 96.05 & -0.67 \\
Qwen3.6-27B & 128K & 96.26 & 2048 & 96.05 & -0.21 \\
Qwen3.6-35B-A3B & 128K & 96.46 & 2048 & 95.80 & -0.66 \\
\bottomrule
\end{tabular}
\end{table}

The important conclusion is not that top-$k=2048$ is always the optimal operating point. In several per-query oracle slices, smaller budgets remain close to dense: for example, Qwen3.5-27B stays within 0.56 points of dense at 128K even with top-$k=256$, and Qwen3.6-27B stays within 0.99 points at 128K with top-$k=128$. These results support the oracle-first hypothesis for the evaluated tasks: dense attention mass often identifies an oracle sparse support that preserves dense behavior.

Table~\ref{tab:qwen35-9b-oracle-prior} adds an end-to-end oracle-prior sweep for Qwen3.5-9B. We report these rows as an attention-mass oracle because the support is chosen directly from the same-layer aggregated dense attention distribution; no distilled indexer weights are used for these rows. The sweep covers all 13 RULER task types with 50 samples per task at each length, for 3900 samples and approximately 140M tokens in total. Across 4K--100K, the oracle remains within 0.48 points of dense under the configured top-$k$ schedule while the reported sparsity is above 85\%.

\begin{table}[t]
\caption{Qwen3.5-9B RULER-13 attention-mass oracle-prior sweep. Oracle support is selected directly from dense attention mass; no indexer weights are used in this table. The top-$k$ values are the configured power-of-two budgets for this sweep, and sparsity is reported under causal-position accounting.}
\label{tab:qwen35-9b-oracle-prior}
\centering
\small
\begin{tabular}{lrrrrr}
\toprule
Context & top-$k$ & Sparsity & Dense & Oracle & Gap \\
\midrule
4K & 256 & 87.9\% & 97.56 & 97.15 & -0.41 \\
8K & 512 & 87.9\% & 96.82 & 96.50 & -0.32 \\
16K & 1024 & 87.9\% & 96.70 & 96.39 & -0.31 \\
32K & 2048 & 87.9\% & 97.38 & 96.97 & -0.41 \\
64K & 2048 & 93.8\% & 96.34 & 96.82 & +0.48 \\
100K & 2048 & 96.0\% & 96.37 & 96.24 & -0.13 \\
\bottomrule
\end{tabular}
\end{table}

Table~\ref{tab:large-top1k-oracle-prior} records the longest-context rows from a second oracle-prior validation focused on larger Qwen3.5/Qwen3.6 checkpoints. Here the configured budget is top-$k=1024$; at 128K this corresponds to 98.44\% sparsity under causal-position accounting. All four 128K rows remain within 0.63 points of dense. The same validation record also includes top-$k=1024$ rows at 32K and 64K with 93.85\% and 96.90\% sparsity, respectively; the reported gaps across those lengths are also below one point. We keep this table separate from Table~\ref{tab:oracle-longest} because the budget is top-$k=1024$ rather than the top-$k=2048$ high-quality slice.

\begin{table}[t]
\caption{Large-checkpoint 128K oracle-prior validation with top-$k=1024$. The Oracle column is an attention-mass oracle selected from dense attention; no indexer weights are used in this table.}
\label{tab:large-top1k-oracle-prior}
\centering
\small
\begin{tabular}{lrrrrr}
\toprule
Model & Context & Sparsity & Dense & Oracle & Gap \\
\midrule
Qwen3.5-27B & 128K & 98.44\% & 96.46 & 96.62 & +0.16 \\
Qwen3.5-35B-A3B & 128K & 98.44\% & 96.52 & 96.05 & -0.47 \\
Qwen3.6-27B & 128K & 98.44\% & 96.26 & 95.85 & -0.41 \\
Qwen3.6-35B-A3B & 128K & 98.44\% & 96.46 & 95.83 & -0.63 \\
\bottomrule
\end{tabular}
\end{table}

\subsection{Support-Sharing and Budget Sensitivity}

The same oracle summaries also show why systems realization matters. Table~\ref{tab:sel64-budget} selects representative selection-block-shared $b_{\mathrm{sel}}=64$ oracle results. In the high-budget region, selection-block-shared oracle support can approach dense quality, but the required budget increases with context length and varies by model. A repeated pattern across Qwen3.5/Qwen3.6 27B and 35B-A3B is that $b_{\mathrm{sel}}=64$ is usually viable around 32K with top-$k=512$, around 64K with top-$k=1024$, and around 128K with top-$k=2048$. When the $b_{\mathrm{sel}}=64$ oracle budget is pushed below this region, retrieval-heavy tasks such as CWE and multi-key NIAH can collapse.

\begin{table}[t]
\caption{Representative selection-block-shared oracle behavior at 128K. High-budget $b_{\mathrm{sel}}=64$ is much closer to dense than the low-budget cliff setting, but it is not uniformly lossless.}
\label{tab:sel64-budget}
\centering
\small
\begin{tabular}{lrrrrrr}
\toprule
Model & Context & Dense & High oracle $k$ & High oracle & Cliff $k$ & Cliff oracle \\
\midrule
Qwen3.5-27B & 128K & 96.46 & 2048 & 96.62 & 256 & 79.44 \\
Qwen3.5-35B-A3B & 128K & 96.72 & 2048 & 95.18 & 256 & 80.36 \\
Qwen3.6-27B & 128K & 96.26 & 2048 & 95.64 & 256 & 79.82 \\
Qwen3.6-35B-A3B & 128K & 96.46 & 2048 & 95.12 & 256 & 79.55 \\
\bottomrule
\end{tabular}
\end{table}

This is the main realization-gap observation in the oracle study. The attention-mass oracle can be close to dense, while the $b_{\mathrm{sel}}=64$ sweep shows that support sharing is viable only when the budget is large enough for the context length and task mix. This observation motivates an inference-time budget schedule and suggests that a candidate sparse-prefill system should consider adapting support granularity or top-$k$ rather than using a single fixed setting across all lengths. In this sense, the oracle is a design tool: it does not merely report that sparse attention is possible, but also indicates which runtime compromises appear plausible under the evaluated settings.

\subsection{Interpretation}

The oracle changes the interpretation of sparse-prefill experiments and the order in which design choices should be made. If per-query oracle sparse attention is near dense, then the chosen top-$k$ and head-averaged token support are plausible for that task slice. If $b_{\mathrm{sel}}=64$ fails below a length-dependent budget, then the problem is not merely sparse budget but support granularity and runtime amortization. A distilled indexer is therefore evaluated against two references: the per-query oracle as an attention-mass feasibility target and the selection-block-shared oracle as a systems-realistic target. These references determine what the indexer should predict, how much support it should keep, and where runtime sharing becomes too coarse.

\section{From Oracle to a Head-Collapsed Indexer}

\subsection{Design Principle}

After the oracle study has established the feasible support family and the realization gap, the next step is to turn that reference into an inference-time indexer. The oracle suggests a conservative design principle:
\begin{quote}
Predict an attention-mass support that approximates dense behavior, and change as little else as possible.
\end{quote}
Accordingly, the original model is kept frozen. The sparse module is an auxiliary indexer trained after pretraining. It consumes hidden states from the frozen model and predicts token importance for sparse top-$k$ selection.

This principle is a direct consequence of the oracle study. The oracle tells us the teacher distribution to imitate, the length-dependent budget region to operate in, and the support granularity that balances quality with runtime amortization. The head-collapsed indexer is therefore an implementation of the oracle guidance under serving constraints: it tries to recover the oracle-selected support without changing the language model's learned computation.

\subsection{KL Distillation from Dense Attention}

The student indexer produces a score $z_{\ell,t}(s)$ for each candidate key position. It is trained to match the dense teacher token-importance distribution $P_{\ell,t}(s)$ through KL divergence:
\[
\mathcal{L}_{\mathrm{KL}} =
\frac{1}{|\mathcal{Q}|}
\sum_{(\ell,t) \in \mathcal{Q}}
\mathrm{KL}\left(
P_{\ell,t} \,\middle\|\, \mathrm{softmax}(z_{\ell,t})
\right),
\]
where the softmax is over valid causal key positions $s\leq t$, and $\mathcal{Q}$ denotes a streamed set of supervised layer/query rows drawn from the internal distillation corpus. Chunked training avoids materializing all attention rows across all layers at once. This release specifies the objective and reported evaluation budgets, but it does not yet expose the exact sampling distribution over layers and positions, the number of supervised rows, or the checkpoint-selection rule as public reproducibility metadata.
All distilled-indexer results reported in this version use this KL distillation objective; we do not mix in additional hit-rate or rank losses.

\subsection{Head-Collapsed Token-Importance Scoring}

A head-expanded auxiliary indexer would maintain separate scoring channels per query head or KV group. The head-averaged training target instead asks the student to predict the shared token-importance distribution $P_{\ell,t}(s)$. The auxiliary scoring space is therefore collapsed across heads:
\[
q_{\ell,t} = \bar{W}_Q x_{\ell,t},
\qquad
k_{\ell,s}^{\mathrm{idx}} = \mathrm{LN}(\bar{W}_K x_{\ell,s}),
\]
followed by positional encoding and scaled similarity:
\[
z_{\ell,t,s}
= \mathrm{ReLU}
\left(
\frac{
\left\langle
\mathrm{RoPE}(q_{\ell,t}), \mathrm{RoPE}(k_{\ell,s}^{\mathrm{idx}})
\right\rangle
}{\sqrt{d_{\mathrm{idx}}}}
\right).
\]
This reduces indexer-side projection activations and removes the need to carry an explicit auxiliary head dimension through scoring and top-$k$ selection. Importantly, this is not backbone head pruning. The original model and its KV cache remain unchanged; the compression applies only to the auxiliary support predictor.

The implementation uses a single sparse index head and considers index dimensions $d_{\mathrm{idx}}\in\{256,128,64\}$. For Qwen3.5, several reported rows use $d_{\mathrm{idx}}=256$, which equals the backbone attention head dimension rather than being smaller. The efficiency argument therefore does not come only from a lower projection dimension; it also comes from removing the auxiliary head dimension, sharing support where the runtime permits it, fusing scoring with top-$k$, and reading only selected values inside sparse attention. This is consistent with the oracle argument: the auxiliary module should predict support rankings, not reproduce full multi-head attention computation. This release does not use these rows as a controlled ablation over $d_{\mathrm{idx}}$.

The head-collapsed scorer should therefore be read as a systems-oriented design choice rather than an ablation-proven optimum. This release evaluates the resulting sparse-prefill path, but it does not isolate the effect of head collapse relative to a head-expanded auxiliary indexer under matched teacher data, training budget, and inference settings.

The ReLU clipping and key-side normalization are implementation choices for stabilizing the scorer and making top-$k$ selection depend on positively aligned keys. The oracle-guided claim does not require this exact scoring function; what matters is whether the distilled ranker recovers the teacher support under a serving-oriented cost.

\subsection{Indexer Cost Model}

A naive inference-time indexer that scores every query-key pair still has $O(S^2d_{\mathrm{idx}})$ scoring work per layer. The indexer is useful only if this work is cheaper than dense attention and can be amortized by systems choices: $d_{\mathrm{idx}}$ can be equal to or smaller than the attention head dimension, the auxiliary head dimension is removed, support can be shared across selection blocks or chunks, top-$k$ can be fused with scoring, and sparse attention avoids loading most values. Therefore, the paper evaluates end-to-end TTFT rather than claiming savings from asymptotic attention FLOPs alone.

For a selection block size $b_{\mathrm{sel}}$, a rough scoring cost per layer is $O((S/b_{\mathrm{sel}})\cdot S\cdot d_{\mathrm{idx}})$ for selection-block-shared support before kernel-specific fusion, while sparse value aggregation costs $O(S\cdot k\cdot d_h)$ at token granularity. This makes the realized speedup sensitive to $b_{\mathrm{sel}}$, $k$, $d_{\mathrm{idx}}$, memory traffic, and top-$k$ overhead.

The serving path executes the sparse full-attention layer in five stages: project hidden states into the indexer space, score candidate keys for each query or query block, select top-$k$ indices, reconstruct K/V from the paged KV cache when the chunked path requires contiguous tensors, and run sparse attention that gathers the selected support inside the attention kernel. In the reported chunked path, full K/V tensors may be reconstructed before sparse attention; the implementation is not limited to reconstructing only the selected KV blocks. The indexer also does not remove all $S^2$-like work automatically; speedup comes from removing the auxiliary head dimension, possible support sharing, fused top-$k$, and avoiding dense value reads in the sparse attention kernel. This release reports end-to-end TTFT and one submodule profile, while a complete kernel-by-kernel cost breakdown remains future work.

\begin{center}
\fbox{\begin{minipage}{0.94\linewidth}
\small
\textbf{Algorithm 2: Oracle-distilled indexer sparse prefill path.}
\begin{enumerate}
\item Train only the auxiliary indexer against the oracle distribution $P_{\ell,t}$ using the KL objective above; keep the backbone frozen.
\item At prefill time, detach hidden states and project them with a single head-collapsed indexer projection to obtain $Q_{\mathrm{idx}},K_{\mathrm{idx}}\in\mathbb{R}^{B\times S\times d_{\mathrm{idx}}}$.
\item Apply key-side normalization, RoPE, and the query scale used by the indexer scorer.
\item Score candidates with $Q_{\mathrm{idx}}K_{\mathrm{idx}}^\top$; in the fused path, apply ReLU and max-aggregate scores over each $b_{\mathrm{sel}}$ query block.
\item Apply causal masking, select top-$k$ token indices, replicate the shared support across KV heads, and run sparse GQA prefill on the original $Q,K,V$ tensors.
\item Keep decode dense in the reported quality runs.
\end{enumerate}
\end{minipage}}
\end{center}

\subsection{Inference-Time Budget Control}

The oracle study also motivates inference-time budget control. The appropriate top-$k$ depends on sequence length, hardware, latency target, and task. Once the indexer is trained to estimate token importance, the same scorer can be used with different budgets at inference time. This makes the oracle more than a training teacher: oracle sweeps suggest an empirical budget schedule, and the indexer is designed to expose that schedule as a quality-latency control without retraining the backbone or the indexer.

\section{System Realization and Sparse Evidence Types}

This section connects the oracle-guided indexer to the serving paths evaluated in the report. Its role is not to introduce a separate system claim before the results; rather, it defines the sparse evidence type, which indexer weights are loaded, and how per-query versus fused selection-block-shared realizations create the realization gap measured later. This prevents oracle sparse rows, distilled-indexer rows, and untrained-indexer stress rows from being read as the same deployed configuration.

\subsection{Three Sparse Evidence Types and Indexer Weights}

The report uses exactly three sparse evidence types. \emph{Oracle sparse} rows do not load indexer weights: support is computed from dense attention mass and then sparse attention is recomputed on that support. These rows test whether a sparse support family is accurate enough under the evaluated budget and granularity. \emph{Distilled-indexer sparse} rows load separately KL-distilled Qwen3.5-0.8B or Qwen3.5-9B indexer weights while keeping the backbone frozen; these are the rows that connect oracle-guided support prediction to output quality and TTFT. \emph{Untrained-indexer stress} rows load undistilled, randomly initialized indexer weights only to exercise the sparse serving path; they measure runtime headroom and are not output-quality evidence.

\begin{table}[t]
\caption{Sparse evidence type and indexer-weight status used by each result group. Untrained-indexer stress rows measure sparse serving headroom and are not output-quality evidence.}
\label{tab:sparse-evidence-types}
\centering
\small
\setlength{\tabcolsep}{4pt}
\begin{tabular}{@{}L{0.22\linewidth}L{0.20\linewidth}L{0.29\linewidth}L{0.18\linewidth}@{}}
\toprule
Result group & Sparse evidence type & Indexer weights / support source & Evidence role \\
\midrule
Oracle tables & oracle sparse & no indexer weights; support is computed from dense attention & quality feasibility under sparse budgets \\
Qwen3.5 quality rows & distilled-indexer sparse & distilled 0.8B and 9B indexers; frozen backbone & output quality with dense decode \\
Qwen3.5-0.8B TTFT & distilled-indexer sparse & distilled 0.8B serving indexer & TTFT / E2E performance for this path \\
Qwen3.5-9B TTFT & distilled-indexer sparse & distilled 9B serving indexer & TTFT / E2E performance for this path \\
Other stress rows & untrained-indexer sparse & undistilled random-init indexer weights & sparse-runtime headroom only \\
\bottomrule
\end{tabular}
\end{table}

\subsection{Per-Query Realization}

The per-query realization selects one top-$k$ support set per query position. It most closely matches the per-query oracle reference and is useful for measuring indexer quality with minimal query sharing. Its limitation is systems cost: selecting and applying different supports for every query is harder to fuse and amortize in current serving implementations.

\subsection{Fused Selection-Block-Shared Realization}

The fused realization selects a support set for a block of $b_{\mathrm{sel}}$ queries and shares it across the block, optionally expanding it across KV groups. This design is intended to improve amortization in indexer scoring, top-$k$ selection, and sparse attention kernels. However, it introduces a realization gap when queries in the same block require different fine-grained evidence. In the reported chunked-prefill rows, 4K or 8K denotes \texttt{max\_num\_batched\_tokens} rather than the model context length.

\subsection{End-to-End TTFT}

\begin{figure}[htbp]
    \centering
    \includegraphics[width=0.9\textwidth]{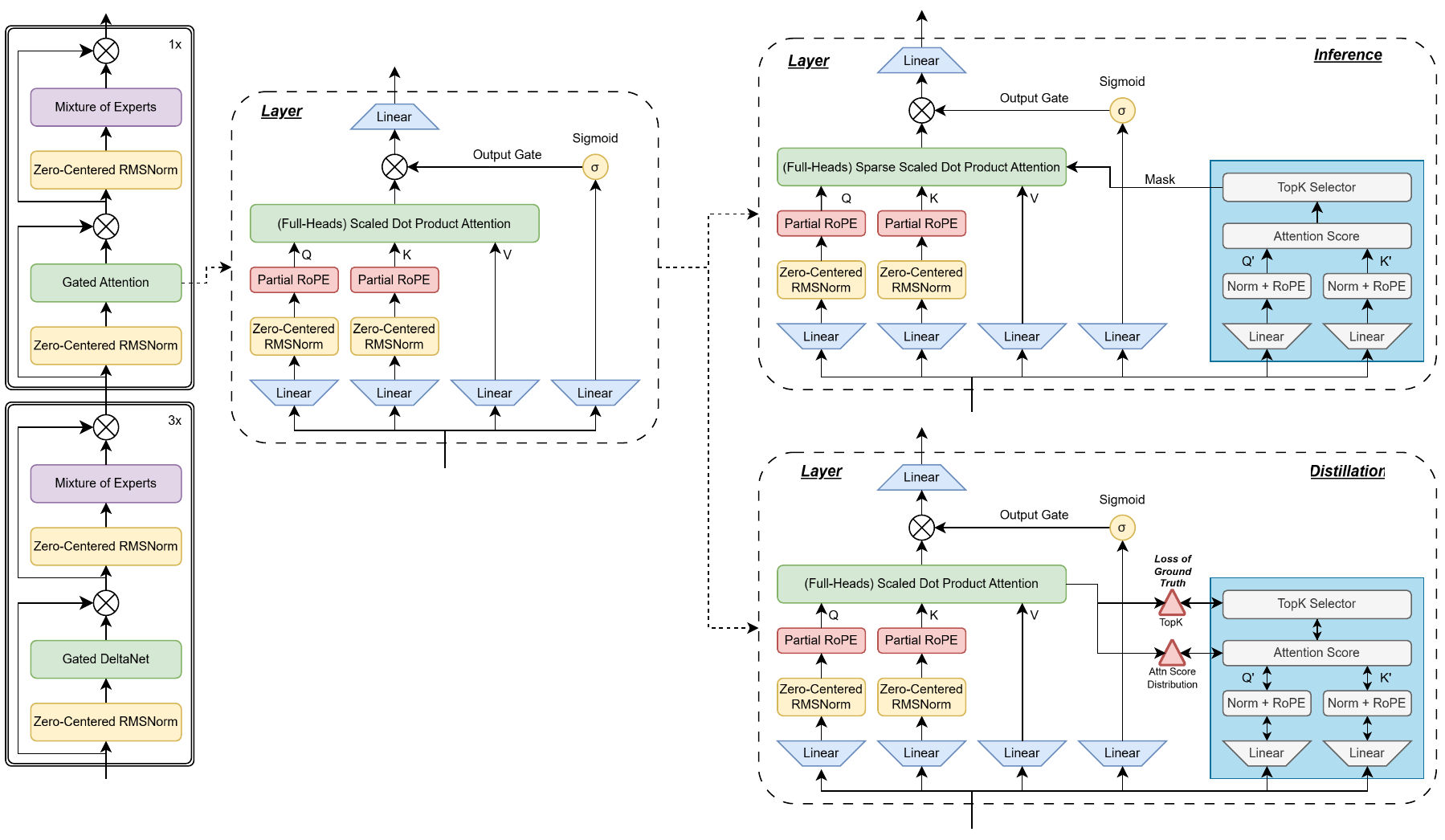}
    \caption{Sparse GQA implementation path for Qwen3.5-style full-attention layers, including the auxiliary indexer scorer, top-$k$ selection, sparse full-attention prefill path, and dense decode boundary.}
    \label{fig:sparse-gqa-implementation}
\end{figure}

Sparse prefill should be evaluated end to end. Attention FLOPs alone are not enough because real TTFT includes embedding, normalization, MLP layers, KV-cache movement, kernel overhead, top-$k$ selection, and serving orchestration. Therefore, the primary systems metric in this prefill-focused release is TTFT under online serving conditions, reported alongside dense-quality comparisons. Figure~\ref{fig:sparse-gqa-implementation} sketches the implementation path used for Qwen3.5-style full-attention layers.

The deployment measurements use our sparse serving implementation. In distilled-indexer quality runs, the backbone is fully frozen: indexer-selected sparse attention is used during prefill, decode remains dense, and only the auxiliary indexer weights are loaded. Decode is therefore not an acceleration target in these runs: autoregressive tokens still attend densely over the cache, so the method does not reduce decode attention work, decode-time KV-cache traffic, or KV-cache storage. This computational boundary does not make generation independent of the sparse prefill stage. The prefix states and cached activations handed to dense decode are produced by sparse-prefill computation, so the decoder continues from a different prompt representation than the dense baseline would have produced. The quality tables therefore evaluate the full sparse-prefill followed by dense-decode generation path, rather than a decode-only intervention.

The Qwen3.5-0.8B TTFT rows are evaluated on a single NPU 64G system with vLLM 0.18.0 online serving and use the distilled 0.8B indexer path. Their dense baseline is the vLLM-Ascend FlashAttention backend registered for dense attention on that platform~\citep{vllm2026attention}. The Qwen3.5-9B TTFT rows are separate GPU 80G vLLM 0.20.0 chunked-prefill measurements and use the distilled 9B serving indexer; their dense baseline is the FlashAttention-2 path in that serving stack. The serving versions therefore follow their measurement stacks, and dense/sparse times should be compared within a row rather than across vLLM versions or hardware classes. Additional stress rows with untrained indexer weights are kept only to exercise the same sparse serving path and reveal runtime headroom.

The TTFT protocol uses one warmup request and ten timed requests per prompt, reports median TTFT, uses the chat prompt template, disables prefix caching, and sets batch/concurrency to 1. This release does not report p95 latency, standard deviation, tokenizer-time accounting, or raw serving logs, so the TTFT rows should be read as representative first-release measurements rather than a full serving benchmark suite.

\section{Results}

\subsection{Evaluation Protocol and Evidence Scope}

This subsection defines the evidence map for the results: which rows are oracle sparse diagnostics, which rows use distilled indexer weights, and which rows use untrained indexer weights for serving stress tests. Table~\ref{tab:protocol} summarizes which evidence supports each claim. Oracle sparse results diagnose sparse-budget feasibility, distilled-indexer sparse results evaluate frozen-backbone sparse prefill with dense decode or TTFT, and untrained-indexer stress rows measure serving speed without making output-quality claims. In the result tables, ``Oracle'' denotes support computed from dense attention mass, ``Distilled'' or ``Indexer'' denotes sparse prefill using distilled real-indexer weights, and ``random-init'' denotes sparse prefill using undistilled randomly initialized indexer weights. The distilled-indexer quality rows are primarily 16K/32K evaluations, while several distilled-indexer and stress-test serving rows measure longer contexts up to 256K. This report therefore does not claim a fully matched dense--oracle--distilled quality-latency frontier. The distilled-indexer quality rows, the distilled-indexer TTFT rows, and the untrained-indexer stress rows should be read as complementary evidence blocks rather than a single claim that the exact same deployed configuration both preserves quality and achieves the largest reported TTFT speedup at every reported length.

We use ``near dense'' narrowly for oracle sparse rows within 1 percentage point of dense on the reported task average. We use ``quality preservation'' for distilled-indexer rows to mean no material aggregate regression under the reported evaluation protocol; it does not imply statistically significant improvement over dense or losslessness on every task cell. Untrained-indexer stress rows are never used as output-quality evidence.

\paragraph{Reproducibility status.}
This first release reports consolidated internal benchmark summaries and serving measurements. Code, distilled indexer checkpoints, raw prompts, raw predictions, evaluator revision hashes, and serving logs are not yet included in the release package. We therefore present the results as first-release evidence with explicit scope boundaries. Future versions will either release these artifacts where possible or provide a table mapping reported cells to model revisions, prompts, scoring scripts, and command lines.

Within each reported evidence block, dense attention is the matched baseline. Prior sparse-prefill and indexer-based systems are discussed as related work, but matched empirical comparisons to MInference, NSA, DSA, Quest, or DuoAttention require aligned models, context lengths, budgets, serving backends, and quality metrics, and are left to later versions.

\begin{table}[!htbp]
\caption{Evaluation protocol and evidence boundaries.}
\label{tab:protocol}
\centering
\small
\setlength{\tabcolsep}{4pt}
\begin{tabular}{@{}L{0.17\linewidth}L{0.22\linewidth}L{0.27\linewidth}L{0.23\linewidth}@{}}
\toprule
Evidence & Models & Data / tasks & Setting \\
\midrule
Oracle quality & Qwen3-8B; Qwen3.5-9B; Qwen3.5/Qwen3.6 27B and 35B-A3B & Non-public RULER-style 13-task averages; Qwen3.5-9B RULER-13 sweep uses 50 samples/task/length & 32K for Qwen3-8B; 4K--100K for Qwen3.5-9B oracle-prior sweep with configured top-$k$ schedule; 32K/64K/128K for larger models; per-query $b_{\mathrm{sel}}=1$, selection-block-shared $b_{\mathrm{sel}}=64$, and oracle-prior rows \\
Distilled-indexer quality & Qwen3.5-0.8B; Qwen3.5-9B & Internal long-context distillation corpus; held-out validation mixture with 1197 samples over RULER, VideoMME, MMLongBench-Doc, and LongBench-v2 & Distilled-indexer quality path; 16K/32K rows reported here; indexer sparse prefill, dense decode, frozen backbone; reported budget schedule uses top-$k=512$ for 4K--16K and top-$k=1024$ for 32K \\
Distilled-indexer TTFT & Qwen3.5-0.8B; Qwen3.5-9B & Online serving TTFT measurements & NPU 64G vLLM 0.18 full prefill with the distilled 0.8B indexer; GPU 80G vLLM 0.20 chunked prefill with the distilled 9B indexer and FA2 dense baseline; not paired with same-context quality rows beyond the reported 16K/32K quality evaluations \\
Untrained-indexer stress TTFT & Qwen3-0.6B; Qwen3-8B; Qwen3.6-27B; Qwen3.6-35B-A3B & Online serving / serving-only TTFT measurements & Undistilled randomly initialized indexer weights; runtime-headroom evidence only, not output-quality evidence \\
\bottomrule
\end{tabular}
\end{table}
\FloatBarrier

\subsection{Finding 1: Evaluated GQA Oracle Slices are Sparse-Reducible}

The primary result is the oracle feasibility result in Tables~\ref{tab:oracle-longest}, \ref{tab:qwen35-9b-oracle-prior}, and~\ref{tab:large-top1k-oracle-prior}. At the longest evaluated context per model, the per-query attention-mass oracle remains within 1 percentage point of dense for all selected Qwen-family GQA rows. The Qwen3.5-9B RULER-13 sweep further shows that this is not only a single-length artifact: from 4K to 100K, the oracle gaps remain between -0.41 and +0.48 points under a configured top-$k$ schedule of 256--2048 while sparsity ranges from 87.9\% to 96.0\%. Because these results do not depend on any indexer weights, they isolate the budget and support-granularity question: for these long-context retrieval-heavy slices, head-averaged oracle sparse support selected by dense attention mass is often sufficient to preserve task-level behavior. This result motivates the paper's design path for GQA sparse prefill: learn the oracle support with a distilled indexer, keep the backbone fixed, and spend systems effort on reducing indexer and realization gaps.

\subsection{Finding 2: Support Sharing Requires Length-Aware Budgets}

The $b_{\mathrm{sel}}=64$ oracle results in Table~\ref{tab:sel64-budget} show that sparse reducibility alone is not enough. A systems-friendly oracle support-sharing policy is much closer to dense at 128K when given top-$k=2048$, but can collapse at top-$k=256$. This gap is especially visible on tasks requiring fine-grained retrieval of multiple needles or exact extracted tokens. These results motivate length-aware, and potentially task-aware, budget control; a single fixed $b_{\mathrm{sel}}=64$ budget is too brittle in the reported 128K retrieval-heavy rows.

\subsection{Finding 3: Distilled Indexers Show No Material Aggregate Degradation}

The distilled-indexer study uses separately distilled Qwen3.5-0.8B and Qwen3.5-9B indexers. Both keep the backbone frozen, load only the distilled indexer, use indexer-selected sparse attention during prefill, and retain dense attention during decode. The distillation data are an internal phase-A long-context corpus with 2K--32K supervision windows. The held-out validation mixture contains 1197 unique samples across RULER/RULER-generated~\citep{hsieh2024ruler}, VideoMME~\citep{fu2024videomme}, MMLongBench-Doc~\citep{ma2024mmlongbenchdoc}, and LongBench-v2~\citep{bai2024longbenchv2}. The reported Table~\ref{tab:distilled-quality} rows use the 16K and 32K slices, totaling 693 samples. Unless otherwise stated, the distilled-indexer sparse budget is top-$k=512$ for 4K--16K and top-$k=1024$ for 32K.

Quality evaluation uses the model chat template with thinking disabled. Decoding follows the default generation configuration used in these runs, including temperature 0.6, sampling top-$k=20$, and top-$p=0.95$. Table~\ref{tab:distilled-quality} reports benchmark scores on a 0--100 display scale while retaining each benchmark's native scoring rule; therefore cross-benchmark rows should not be interpreted as a single homogeneous accuracy metric. The average row is a displayed-row macro average rather than a statistically weighted benchmark aggregate. Positive distilled-minus-dense gaps are interpreted as quality preservation within the reported evaluation resolution, not as evidence that sparse prefill improves the base model.

\begin{table}[!htbp]
\caption{Held-out validation sample distribution. The 16K and 32K rows are the slices reported in Table~\ref{tab:distilled-quality}.}
\label{tab:validation-samples}
\centering
\small
\begin{tabular}{lrrrrr}
\toprule
Context & LongBench-v2 & MMLongBench-Doc & VideoMME & RULER / generated & Total \\
\midrule
4K & -- & 13 & 96 & 91 & 200 \\
8K & -- & 100 & 100 & 104 & 304 \\
16K & 38 & 100 & 100 & 100 & 338 \\
32K & 51 & 100 & 100 & 104 & 355 \\
\midrule
Total & 89 & 313 & 396 & 399 & 1197 \\
\bottomrule
\end{tabular}
\end{table}
\FloatBarrier

This release specifies sample counts, chat-template use, thinking-mode setting, and the default sampling parameters used in the reported runs. It does not include raw predictions, evaluator revision hashes, bootstrap confidence intervals, or repeated-run variance. For this reason, positive distilled-minus-dense differences in Table~\ref{tab:distilled-quality} are described as quality preservation rather than quality improvement.

\begin{table}[!htbp]
\caption{Held-out quality of distilled indexers with sparse prefill and dense decode. Gap is Distilled minus Dense, in percentage points. The average row is a displayed-row macro average; positive gaps are treated as quality preservation under the reported protocol, not as improvement over dense attention.}
\label{tab:distilled-quality}
\centering
\small
\begin{tabular}{llrrrrrr}
\toprule
Benchmark & Context & \multicolumn{3}{c}{Qwen3.5-0.8B} & \multicolumn{3}{c}{Qwen3.5-9B} \\
\cmidrule(lr){3-5}\cmidrule(lr){6-8}
 & & Dense & Distilled & Gap & Dense & Distilled & Gap \\
\midrule
RULER & 16K & 70.47 & 68.33 & -2.14 & 90.00 & 90.00 & +0.00 \\
RULER & 32K & 91.25 & 89.52 & -1.73 & 95.83 & 95.83 & +0.00 \\
VideoMME & 16K & 51.00 & 55.00 & +4.00 & 65.00 & 66.00 & +1.00 \\
VideoMME & 32K & 51.00 & 51.00 & +0.00 & 68.00 & 70.00 & +2.00 \\
MMLongBench-Doc & 16K & 40.00 & 42.00 & +2.00 & 47.14 & 48.57 & +1.43 \\
MMLongBench-Doc & 32K & 34.00 & 35.00 & +1.00 & 45.00 & 45.00 & +0.00 \\
LongBench-v2 & 16K & 18.42 & 31.58 & +13.16 & 55.26 & 57.89 & +2.63 \\
LongBench-v2 & 32K & 29.41 & 29.41 & +0.00 & 54.90 & 56.86 & +1.96 \\
\midrule
Macro average & -- & 48.19 & 50.23 & +2.04 & 65.14 & 66.27 & +1.13 \\
\bottomrule
\end{tabular}
\end{table}
\FloatBarrier

Table~\ref{tab:distilled-quality} shows no material aggregate degradation from the distilled indexer within the resolution of the reported evaluation. Since several distilled-minus-dense gaps are positive and no confidence intervals are reported, we treat these results as quality preservation rather than improvement over dense attention. Individual rows still reveal task sensitivity, especially for 0.8B RULER; the result is not a claim of losslessness for every task cell.

\begin{table}[!htbp]
\caption{Independent pure RULER-32K validation using all RULER task types. This table is separate from the 1.2K mixed benchmark in Table~\ref{tab:distilled-quality}. For Qwen3.5-0.8B, the reported GPU 80G run includes both per-query and fused distilled-indexer pipelines over 13 tasks with 100 samples per task; for Qwen3.5-9B, the reported summary does not separate those pipelines.}
\label{tab:ruler32-distilled}
\centering
\small
\setlength{\tabcolsep}{4pt}
\begin{tabular}{@{}lrrrrL{0.35\linewidth}@{}}
\toprule
Model & Dense & Distilled & Fused distilled & Best gap & Setting \\
\midrule
Qwen3.5-0.8B & 90.8 & 90.2 & 85.8 & -0.6 & distilled indexer, top-$k=1024$, Distilled column is per-query \\
Qwen3.5-9B & 95.2 & 95.2 & -- & +0.0 & distilled indexer, top-$k=1024$, per-query/fused split unavailable \\
\bottomrule
\end{tabular}
\end{table}
\FloatBarrier

The independent RULER-32K validation in Table~\ref{tab:ruler32-distilled} is particularly important because it removes ambiguity about benchmark mixture effects. On all RULER task types, the distilled-indexer per-query sparse-prefill path is within 0.6 points of dense for Qwen3.5-0.8B, while the distilled-indexer fused selection-block-shared path exposes a 5.0 point realization gap. The Qwen3.5-9B row is reported as matching dense, but this summary does not separate per-query and fused realizations for that model. The RULER-32K dense baselines in Tables~\ref{tab:distilled-quality} and~\ref{tab:ruler32-distilled} come from different evaluation slices: Table~\ref{tab:distilled-quality} reports the RULER slice inside the 1.2K mixed validation bucket, while Table~\ref{tab:ruler32-distilled} reports a separate pure all-task RULER run with 100 samples per task. Their different dense scores therefore reflect different sample sets rather than contradictory baselines for one matched frontier.

\clearpage

\subsection{Finding 4: Distilled Indexer Serving Shows End-to-End TTFT Headroom}

Table~\ref{tab:serving-ttft} reports end-to-end online-serving TTFT for distilled-indexer sparse prefill. The Qwen3.5-0.8B rows are full-prefill measurements on one NPU 64G system with vLLM 0.18.0 and use the distilled 0.8B indexer path, while the Qwen3.5-9B rows are chunked-prefill measurements on one GPU 80G system with vLLM 0.20.0 and use the distilled 9B serving indexer. These rows are the main real-indexer systems evidence in this release: they use the same class of distilled support predictor that is evaluated for output quality in Tables~\ref{tab:distilled-quality} and~\ref{tab:ruler32-distilled}. They are still not a fully matched quality-latency frontier because the reported quality rows are primarily 16K/32K, while the longest TTFT rows extend to 256K. The 4K chunk entry denotes \texttt{max\_num\_batched\_tokens}, not a model context limit.

\begin{table}[!htbp]
\caption{End-to-end TTFT of distilled-indexer sparse prefill. Dense and sparse times are seconds; all rows are single-card online-serving measurements. Qwen3.5-0.8B uses vLLM Ascend FlashAttention as the dense baseline; Qwen3.5-9B measurements were collected in milliseconds and are shown in seconds, with FlashAttention-2 as the dense baseline in the GPU 80G vLLM 0.20 chunked-prefill stack.}
\label{tab:serving-ttft}
\centering
\small
\setlength{\tabcolsep}{3pt}
\resizebox{\textwidth}{!}{%
\begin{tabular}{@{}llllllrrrr@{}}
\toprule
Model & Indexer & Hardware & Serving & Dense baseline & Context & top-$k$ & Dense (s) & Sparse (s) & Speedup \\
\midrule
Qwen3.5-0.8B & distilled & NPU 64G & vLLM 0.18 full & Ascend FA & 128K & 1024 & 5.957 & 4.613 & 1.29$\times$ \\
Qwen3.5-0.8B & distilled & NPU 64G & vLLM 0.18 full & Ascend FA & 256K & 1024 & 17.928 & 10.491 & 1.71$\times$ \\
Qwen3.5-9B & distilled & GPU 80G & vLLM 0.20 chunk 4K & FA2 & 32K & 1024 & 2.944 & 2.747 & 1.07$\times$ \\
Qwen3.5-9B & distilled & GPU 80G & vLLM 0.20 chunk 4K & FA2 & 64K & 2048 & 6.863 & 6.024 & 1.14$\times$ \\
Qwen3.5-9B & distilled & GPU 80G & vLLM 0.20 chunk 4K & FA2 & 100K & 2048 & 11.780 & 9.173 & 1.28$\times$ \\
Qwen3.5-9B & distilled & GPU 80G & vLLM 0.20 chunk 4K & FA2 & 128K & 2048 & 17.652 & 12.452 & 1.42$\times$ \\
Qwen3.5-9B & distilled & GPU 80G & vLLM 0.20 chunk 4K & FA2 & 256K & 2048 & 50.915 & 26.322 & 1.93$\times$ \\
\bottomrule
\end{tabular}}
\end{table}
\FloatBarrier

The TTFT trend is consistent with the oracle-guided expectation that serving-time sparse prefill can offer larger reductions as prompt length grows in the reported serving rows. For Qwen3.5-9B, the distilled-indexer path is modestly faster than dense FA2 at 32K, where non-attention layers, serving overheads, and indexer work still remain, but the speedup increases from 1.07$\times$ at 32K to 1.93$\times$ at 256K. The 100K row is included because it aligns with the Qwen3.5-9B oracle-prior sweep length. Taken together, Tables~\ref{tab:distilled-quality}, \ref{tab:ruler32-distilled}, and~\ref{tab:serving-ttft} provide the strongest matched evidence block in this release: distilled Qwen3.5 indexers preserve aggregate quality on the reported validation slices and reduce TTFT on measured NPU and GPU serving paths. A stronger deployment claim would still require the matched quality-latency frontier in Section~\ref{sec:evidence-boundaries}, including same-context 128K/256K quality checks for the distilled 0.8B serving rows and same-context quality checks beyond 32K for the distilled 9B serving rows.

\clearpage

\subsection{Finding 5: Untrained Indexer Stress Tests Show Additional Runtime Headroom}

Table~\ref{tab:latency-only} keeps additional sparse-serving measurements that run the same sparse serving path with untrained indexer weights. These rows are useful for understanding systems scaling across GPU and NPU serving paths, but they are not output-quality results because the indexer weights are undistilled and randomly initialized. Their role is narrower but still important: they show how much TTFT headroom the sparse runtime can expose when the support-selection path is exercised. For checkpoints that also have oracle rows, such as Qwen3-8B and Qwen3.6, the oracle results provide sparse-support feasibility while the untrained-indexer stress rows provide runtime-headroom evidence. These two blocks are complementary, not a proof that the exact stress-test serving configuration preserves quality. Qwen3-0.6B is even narrower in this release: it has NPU runtime and profiling evidence but no reported oracle or distilled-indexer quality result.

\begin{table}[!htbp]
\caption{Additional TTFT stress tests with untrained indexer weights. These rows exercise the sparse serving path and measure runtime scaling, but they do not establish output quality.}
\label{tab:latency-only}
\centering
\small
\begin{tabular}{llllrrrr}
\toprule
Model & Hardware & Indexer weights & Context & top-$k$ & Dense (s) & Sparse (s) & Speedup \\
\midrule
Qwen3-0.6B & NPU 64G & random-init & 64K & 1024 & 4.528 & 2.402 & 1.89$\times$ \\
Qwen3-0.6B & NPU 64G & random-init & 128K & 1024 & 18.018 & 5.233 & 3.44$\times$ \\
Qwen3-8B & GPU 80G & random-init & 32K & 512 & 4.21 & 2.97 & 1.42$\times$ \\
Qwen3-8B & GPU 80G & random-init & 64K & 1024 & 12.31 & 6.72 & 1.83$\times$ \\
Qwen3-8B & GPU 80G & random-init & 128K & 2048 & 39.52 & 16.37 & 2.42$\times$ \\
Qwen3.6-27B & GPU 80G & random-init & 128K & 1024 & 58.241 & 35.498 & 1.64$\times$ \\
Qwen3.6-35B-A3B & GPU 80G & random-init & 128K & 1024 & 16.126 & 7.652 & 2.11$\times$ \\
\bottomrule
\end{tabular}
\end{table}
\FloatBarrier

The stress rows are useful as systems diagnostics because they show sparse-runtime headroom on measured checkpoints and hardware paths that do not yet have distilled-indexer quality validation. The Qwen3-0.6B NPU full-prefill rows show large sparse-runtime headroom at 64K and 128K, reaching 3.44$\times$ at 128K, but they remain runtime-only stress rows. Untrained indexer weights can exercise TTFT/E2E serving paths, but Qwen3-0.6B, Qwen3-8B, and Qwen3.6 require distilled-indexer quality evaluation before they can support the same deployment claim as Qwen3.5.

\subsection{Finding 6: Profiling-Only Qwen3-0.6B Suggests a Candidate Bottleneck}

As a profiling-only stress case, submodule profiling on Qwen3-0.6B illustrates why indexer-side costs matter. This row is not part of the distilled-indexer quality evidence. At 128K, averaged over 28 layers, indexer projection costs only 1.81 ms per layer, or 1.4\% of the sparse pipeline. The dominant stages are the fused indexer at 55.55 ms per layer (42.1\%) and sparse prefill at 64.26 ms per layer (48.7\%). This profiling row helps interpret the Qwen3-0.6B NPU full-prefill speedups in Table~\ref{tab:latency-only}: once sparse attention itself becomes cheaper, candidate scoring and top-$k$ expansion can become candidate systems bottlenecks. This motivates treating support selection as a first-class systems problem rather than a small auxiliary detail.

\subsection{Evidence Boundaries}
\label{sec:evidence-boundaries}

The reported tables support three sparse evidence types with different scopes. The oracle tables show that the evaluated attention-mass oracle sparse supports are close to dense under specified budgets. The distilled-indexer quality tables show no material aggregate degradation for frozen-backbone indexer sparse prefill with dense decode on Qwen3.5-0.8B and Qwen3.5-9B at the reported validation lengths. The distilled-indexer TTFT table shows single-card online-serving speedups for Qwen3.5-0.8B and Qwen3.5-9B with real distilled indexer weights. The additional stress-test table uses untrained indexer weights on Qwen3-0.6B, Qwen3-8B, and Qwen3.6 checkpoints and is a systems diagnostic rather than output-quality evidence.

These tables should not be combined into a deployment frontier. This technical report does not yet provide a single table where dense, per-query oracle sparse, selection-block-shared oracle sparse, distilled per-query indexer, distilled fused indexer, and online TTFT are all measured on the same model, benchmark slice, context length, top-$k$, indexer checkpoint, and serving stack. In particular, the distilled-indexer quality tables are reported at 16K/32K, while the longest distilled-indexer TTFT rows extend to 256K. The reported evidence supports oracle-guided sparse-prefill feasibility, distilled-indexer quality on the reported validation slices, and serving-side systems headroom. For checkpoints without distilled indexer validation, oracle rows and untrained-indexer stress rows should be read as complementary feasibility and runtime evidence only when both are available; Qwen3-0.6B remains runtime/profiling-only in this release. A stricter deployment claim requires the matched frontier.

\section{Analysis and Future Work}

\subsection{Oracle Gap vs. Indexer Gap}

The oracle gap measures the loss of the attention-mass support itself. The indexer gap measures how much additional quality is lost when a distilled indexer replaces oracle support. The oracle tables directly measure oracle gap for per-query and selection-block-shared oracle paths, while the distilled-indexer tables measure end quality after indexer distillation and runtime realization. In this release, the indexer gap is not isolated with support-level diagnostics such as teacher top-$k$ overlap, distilled-indexer recall of oracle support, or validation KL under matched benchmark slices. Those diagnostics are the most direct next step for turning the reported feasibility evidence into a full error attribution study.

\subsection{Budget Sensitivity}

The $b_{\mathrm{sel}}=64$ oracle sweeps show that a fixed budget is risky. The empirical pattern in the reported summaries is that selection-block-shared oracle support needs roughly one larger top-$k$ tier for each 2$\times$ increase in context length: 32K often tolerates top-$k=512$, 64K typically needs top-$k=1024$, and 128K is least risky at top-$k=2048$ among the reported settings. This rule is an empirical starting point rather than a universal law.

\subsection{Support Granularity}

Per-query oracle support is more accurate in the reported rows but harder to amortize in serving. Selection-block-shared oracle support is more amenable to amortized serving but can lose fine-grained retrieval. The $b_{\mathrm{sel}}=64$ cliff examples at 128K show that support sharing can lose sharply when the budget is too low. The key systems challenge is to reduce this realization gap through better grouping, adaptive block sizes, or hybrid support policies.

\subsection{Head-Collapsed vs. Head-Expanded Indexers}

The representation question is whether the head-collapsed scorer preserves distilled-indexer quality while reducing indexer activation cost, scoring overhead, and TTFT relative to a head-expanded auxiliary indexer. This report motivates the head-collapsed design and reports its quality/TTFT behavior, but it does not include a controlled head-collapsed versus head-expanded ablation. A future version should compare trainable parameters, projection activation size, validation KL, oracle-support overlap, task score, per-layer indexer latency, and end-to-end TTFT under the same teacher data.

\subsection{Cross-Family Validation}

The oracle evidence is concentrated on Qwen3/Qwen3.5/Qwen3.6 checkpoints, while the method is formulated for GQA layers rather than for a specific model family. Extending the empirical claim to broader hybrid long-context architectures requires comparable oracle and distilled-indexer studies on at least one non-Qwen family with a different attention layout, such as a Gemma-style local/global GQA model.

\section{Limitations and Scope of This Release}

This first release has several scope boundaries. The empirical evidence is concentrated on Qwen-family GQA checkpoints and retrieval-heavy RULER-style or mixed long-context evaluations; the method is formulated for GQA layers more generally, but cross-family evidence remains future work. The attention-mass top-$k$ oracle is a controlled reference for head-averaged token support, not a deployable acceleration method. The reported sparse path targets prefill rather than autoregressive decode: dense decode still attends over the cache and therefore receives no decode-side attention, KV-traffic, or KV-storage reduction, although it consumes prefix states and cached activations produced by sparse prefill. The distilled-indexer quality evidence is strongest for Qwen3.5-0.8B and Qwen3.5-9B at the reported 16K/32K validation lengths with dense decode, while the longest TTFT measurements are not paired with same-context quality evaluations. The Qwen3.5-9B serving rows now use the distilled serving indexer, but the 64K/100K/128K/256K TTFT rows still do not have same-context quality rows in this release. Qwen3-8B and Qwen3.6 have oracle feasibility evidence and untrained-indexer runtime stress evidence but no distilled-indexer quality result. Qwen3-0.6B has NPU full-prefill latency and profiling-only evidence but no reported oracle or distilled-indexer quality result. Additional untrained-indexer stress rows are systems diagnostics rather than output-quality evidence.

This version also does not yet provide a fully matched dense--oracle--distilled--fused quality-latency frontier, matched comparisons to other sparse-prefill systems, confidence intervals or repeated-run variance, public raw predictions and serving logs, or a kernel-level cost breakdown. The head-collapsed indexer is evaluated as the design used in this report, but not yet isolated against a matched head-expanded auxiliary indexer. Finally, $b_{\mathrm{sel}}=64$ selection-block-shared oracle support can lose sharply on retrieval-heavy tasks when the budget is too low, so candidate deployed realizations should consider length-aware budget control or finer support granularity. These limitations do not change the main role of the report: to establish an oracle-guided evidence framework for sparse prefill and to report first-release evidence for oracle feasibility, distilled-indexer quality, and serving-side TTFT headroom under explicitly scoped settings.

\section{Conclusion}

This paper reframes GQA sparse prefill as an oracle-guided reducibility problem. Instead of beginning with a sparse module and asking whether it works, we first ask how much dense attention is necessary under explicit support constraints. The attention-mass top-$k$ oracle is the organizing primitive of the method: it measures sparse feasibility, defines the distillation target, guides budget and support granularity, and organizes the oracle/indexer/realization gap analysis. The selection-block-shared oracle further indicates that systems-oriented realizations need enough budget and granularity to avoid retrieval cliffs in the reported settings. The current evidence separately supports three claims: oracle sparse supports are often near dense under the reported budgets; distilled Qwen3.5 indexers preserve aggregate quality on the reported validation slices and reduce TTFT on measured NPU/GPU serving paths; and additional untrained-indexer stress rows expose sparse-runtime headroom on the same serving family without validating output quality. As a first-release technical report, the paper provides the oracle-guided GQA sparse-prefill story and the reported Qwen-family evidence, while leaving fully matched quality-latency frontiers, richer statistical reporting, and cross-family validation to future versions.

\end{document}